%% file: root.tex
\documentclass[letterpaper, 10 pt, conference]{ieeeconf}  

\IEEEoverridecommandlockouts                              

\usepackage{amsmath} 
\usepackage{amssymb}  
\usepackage{stackengine}
\stackMath
\usepackage{cite}
\usepackage[ruled,noend]{algorithm2e}
\usepackage{graphicx}
\usepackage{caption}
\usepackage{subcaption}
\usepackage{hyperref}
\usepackage[switch]{lineno}
\usepackage{mathtools}

\makeatletter
\renewenvironment{subarray}[2][c]{%
  \if#1c\vcenter\else\vbox\fi\bgroup
  \Let@ \restore@math@cr \default@tag
  \baselineskip\fontdimen10 \scriptfont\tw@
  \advance\baselineskip\fontdimen12 \scriptfont\tw@
  \lineskip\thr@@\fontdimen8 \scriptfont\thr@@
  \lineskiplimit\lineskip
  \ialign\bgroup\ifx c#2\hfil\fi
    $\m@th\scriptstyle##$\hfil\crcr
}{%
  \crcr\egroup\egroup
}
\makeatother
\renewcommand{\substack}[2][c]{\subarray[#1]{c}#2\endsubarray}

\input{macros.tex}
\usepackage{xcolor}
\usepackage{soul}

\title{Dynamically Switching Human Prediction Models for Efficient Planning}

\author{Arjun Sripathy$^*$, Andreea Bobu$^*$, Daniel S. Brown, and Anca D. Dragan
\thanks{*Indicates equal contribution.
Authors are with EECS at UC Berkeley. 
Research supported by the Air Force Office of Scientific Research (AFOSR),
NSF grant IIS1734633 (SCHooL), and NSF grant IIS1652083 (CAREER).}
}
\begin{document}
\maketitle
\thispagestyle{empty}
\pagestyle{empty}


\begin{abstract}

As environments involving both robots and humans become increasingly common, so does the need to account for people during planning. To plan effectively, robots must be able to respond to and sometimes influence what humans do. This requires a \emph{human model} which predicts future human actions. A simple model may assume the human will continue what they did previously; a more complex one might predict that the human will act optimally, disregarding the robot; whereas an even more complex one might capture the robot's ability to influence the human. These models make different trade-offs between computational time and performance of the resulting robot plan. Using only one model of the human either wastes computational resources or is unable to handle critical situations.
In this work, we give the robot access to a suite of human models and enable it to assess the performance-computation trade-off online.
By estimating how an alternate model could improve human prediction and how that may translate to performance gain, the robot can \textit{dynamically switch human models} whenever the additional computation is justified.
Our experiments in a driving simulator showcase how the robot can achieve performance comparable to always using the best human model, but with greatly reduced computation.



\end{abstract}

\input{1_intro}
\input{2_method}

\input{3_experiments}
\input{4_conclusion}






{
\bibliographystyle{IEEEtran}
\bibliography{IEEEabrv,references}}

\end{document}

%% file: macros.tex
\newcommand{\state}{\ensuremath{x}}
\newcommand{\xtraj}{\ensuremath{\mathbf{x}}}

\newcommand{\control}{u}
\newcommand{\utraj}{\ensuremath{\mathbf{u}}}
\newcommand{\uset}{\ensuremath{\mathcal{U}}}

\newcommand{\model}{\ensuremath{M}}
\newcommand{\modelControl}{\hat\control_H}
\newcommand{\modelTraj}{\hat\utraj_H}
\newcommand{\plan}{\hat\utraj_R}
\newcommand{\planControl}{\hat\control_R}

\newcommand{\modelControlAct}{\bar\control_H}
\newcommand{\modelTrajAct}{\bar\utraj_H}
\newcommand{\planAct}{\bar\utraj_R}
\newcommand{\planControlAct}{\bar\control_R}

%% file: 1_intro.tex
\section{Introduction}
\label{sec:intro}

When robots operate in close proximity to humans, it is crucial that they anticipate what people will do to respond appropriately. 
Such prediction often involves equipping the robot with a model of human behavior~\cite{rudenko2019survey}. 
This model could be physics-based~\cite{barth2008where,schubert2008comparison,helbing1995socialforces}, pattern-based ~\cite{lefevre2015models,alahi2016socialLSTM,Schmerling2018cvae,Pfeiffer2016PredictingAT}, approximate rationality with respect to an objective function~\cite{ziebart2009planning,kretzschmar2016socially,Kretzschmar2014learning,Kuderer2012FeatureBasedPO,bobu2020LESS,Liu2016GoalII,Mainprice2015PredictingHR,kitani2012forecast,Sunberg2017TheVO},
or even two player games~\cite{Sadigh2016PlanningFA,Fisac2017PragmaticPedagogicVA,brown2019machine,FridovichKeil2020EfficientIL}. 

\begin{figure}
\includegraphics[width=0.4\textwidth]{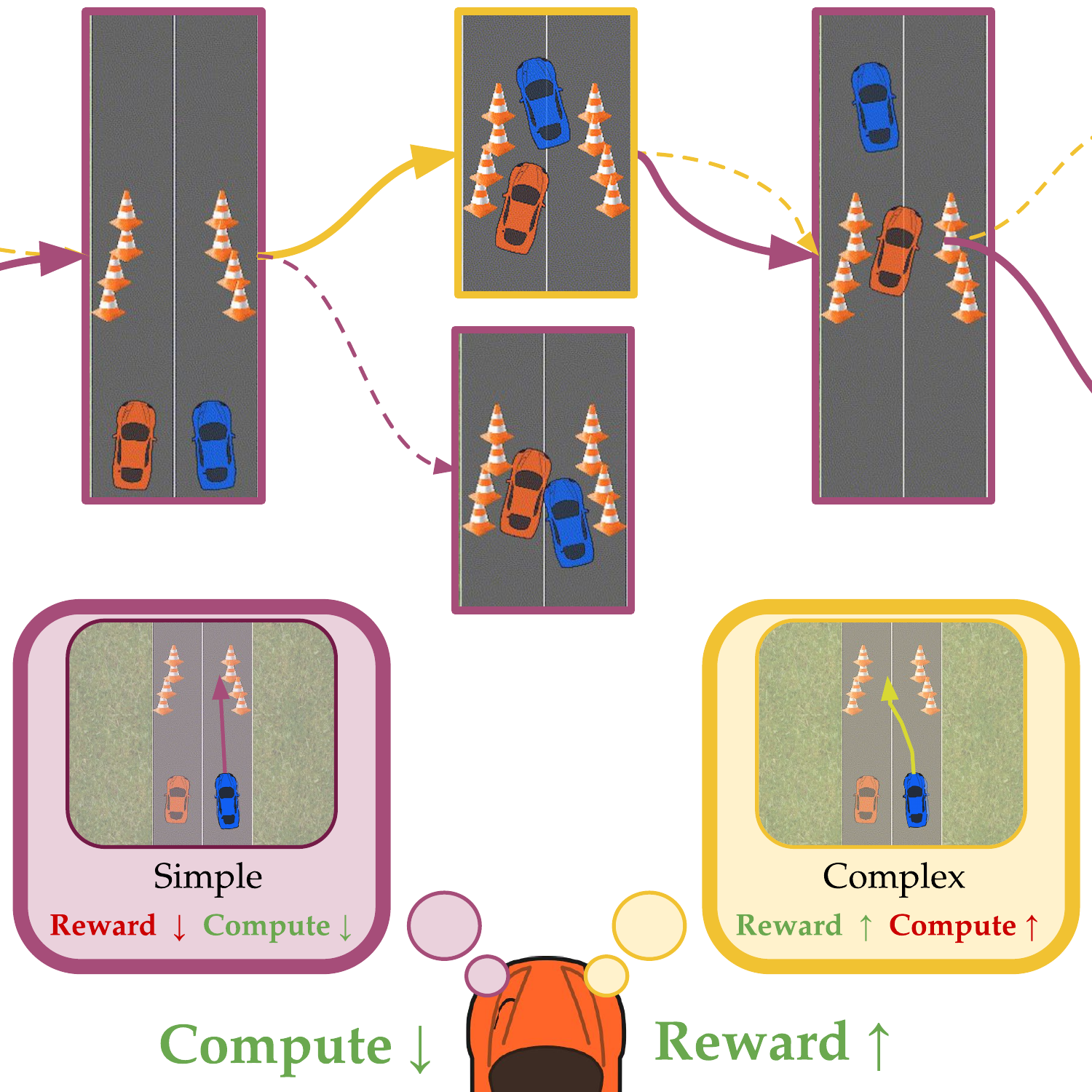}
\centering
\caption{The robot (orange car) can plan with either a complex model (yellow bubble) of the human (blue car) that is more accurate but more expensive or with a simple model (purple bubble) that is not as accurate but cheaper. Our algorithm uses the complex model when a collision is imminent (solid yellow line), but saves computation by switching to the simple one afterwards (solid purple line).}
\label{fig:front_fig}
\end{figure}

What human model a robot should be equipped with depends on the trade-offs it needs to make: some models, like the physics-based ones, are cheap to compute but don't capture human intentions and may be less accurate; others, like two player games, model interactions between the person and the robot, but at a high computational cost.
For systems interacting with people in real time, like autonomous vehicles or mobile robots, compromising on either accuracy or computation is undesirable. For instance, in the driving scenario in Fig. \ref{fig:front_fig}, using the cheap model might pose a safety hazard, while picking the more accurate one may limit planning frequency or strain computational resources needed elsewhere (sensor suite, perception system, routing, etc).

We advocate that the robot should not be stuck with a single model, but instead have the ability to dynamically change which model it is using as the interaction progresses. In Fig. \ref{fig:front_fig}, the robot starts off with the cheap model. Anticipating a potential collision, it switches to a more complex model, reverting back once the critical maneuver is complete.

The idea of using multiple predictive models is not new; most works, like interactive multiple model filtering~\cite{li2005multiplemodel,Pool2017UsingRT,pentland1999modeling,lasota2017multiple} or other ways of combining models~\cite{lasota2017multiple} are focused on improving accuracy by leveraging complementary strengths of different models. In contrast, we are focused on the setting where we have complex but accurate models (which could be mixtures themselves), and cheap but less accurate ones. In such settings, if it weren't for computational costs, we would use the complex models all the time. The question becomes: when is the performance gain worth the extra computation?

We could plan with the complex model and measure the performance gain, but doing so defeats the purpose of saving computation. 
To avoid this, prior work proposed training a model to predict which agents ``influence'' or ``affect'' the planner, and prioritizing computation for them~\cite{refaat2019prioritization}. 
This approach does not estimate the performance-computation trade off, but heuristically assumes that agents who influence the planner are conducive to high gain. 
However, there is a fundamental difference between needing to consider an agent at all, and estimating the performance gain between predicting their behavior using a cheap versus a complex model.
For example, in Fig. \ref{fig:front_fig}, a cheap model is sufficient in the leftmost and rightmost frames where a critical maneuver is not necessary, despite the human influencing the planner.  

Instead of heuristically allocating computation, we employ efficient, online estimation for how an alternate human model could change robot performance. This enables switching to the model which best trades off reward and computation in real time.
In Fig. \ref{fig:front_fig}, a car using our switching methodology is able to achieve a behavior similar to the one produced with the most accurate model, but at a computational cost closer to that of the cheaper model.

This paper makes three key contributions: (1) a formalism for the robot's decision process that optimally trades off between computation and accuracy across multiple predictive models, (2) an approximate solution that solves this decision process online, and (3) a comparative analysis of our model switching algorithm in a human-autonomous car system. Together, these contributions give robots the autonomy to decide in real-time what predictive human models are most appropriate to use in different interactive scenarios. Code and videos are made available at \href{https://arjunsripathy.github.io/model_switching}{arjunsripathy.github.io/model\_switching}

%% file: 2_method.tex
\section{Method}
\label{sec:method}

We focus on a system consisting of a robot $R$ interacting in an environment with other human agents $H$. 
The robot's goal is to plan around the humans in a manner that is most effective while minimizing computational time. 
We present our theory for a general single human, single robot setting where the other agents’ behavior is known, although our method can easily be extended by running it separately for each human.
We use the running example of an autonomous car sharing the road with a human driver to illustrate the proposed approach and demonstrate the utility of our method.

\subsection{Problem Statement}

We model the system as a fully observable dynamical system in which one agent's controls potentially impact the other's. As such, let the state $\state \in \mathcal{X}$ include the positions and velocities of both agents, where the human action $\control_H\in\uset_H$ and robot action $\control_R\in\uset_R$ each can affect the next state in a combined dynamics model: $\state^{t+1} = f(\state^t, \control_R^t, \control_H^t)$.

Let $\xtraj = [\state^1, \hdots, \state^N]$ be a finite horizon $N$ state sequence, $\utraj_R = [\control_R^1, \hdots, \control_R^N]$ the robot's continuous control inputs, and $\utraj_H = [\control_H^1, \hdots, \control_H^N]$ the human's. The robot optimizes its controls $\utraj_R$ according to a reward function that depends on the joint sequence of states and controls: $R_R(\state^0,\utraj_R,\utraj_H) = \sum_{\tau=1}^N r_R(\state^\tau, \control_R^\tau, \control_H^\tau)$, where $\state^0$ is the starting state and each state thereafter is obtained via the dynamics model from the previous robot and human controls. 
The person chooses their action at time $t$, $\control_H^t$, according to an internal policy $\pi_H(\state^t, \control_R^t)$, which, when applied at every state $\state^t$, results in $\utraj_H$.
We let $\utraj_R$ and $\utraj_H$ represent the true executed robot and human controls, respectively.

Our system is a Markov Decision Process (MDP) with states $\mathcal{X}$, actions $\uset_R$, transition function $f(\state, \control_R, \control_H)$, and reward $r_R(\state,\control_R,\control_H)$.
Since the robot does not know $\pi_H$ (that would require access to the human's brain), it uses a \emph{human model} $\model:\mathcal{X}\times\uset_R^N\rightarrow\uset_H^N$ to make a \emph{prediction} of the human controls $\modelTrajAct$. The robot then seeks a \emph{plan} $\planAct^0(\model)\coloneqq\planAct(\state^0, \model)$ that maximizes the MDP reward $R_R$ based on
the predicted $\modelTrajAct^0(\planAct, \model)\coloneqq\modelTrajAct(\state^0, \planAct, \model)$.

Unfortunately, this type of \textit{offline planning} doesn't consider modeling errors introduced by imperfect models $\model$. Hence, it is more common for the robot to perform \textit{online planning} at every time step $t$ to obtain more accurate plans $\planAct^t(\model)$. For computational efficiency, we follow~\cite{Sadigh2016PlanningFA} and use Model Predictive Control (MPC)~\cite{garcia1989MPC} with a finite horizon $K<N$, where at each time step $t$ the robot optimizes its controls to maximize the cumulative reward: 
%
\begin{equation}
    \planAct^t(\model) =  \arg\max_{\utraj_R} R_R(\state^t, \utraj_R, \modelTrajAct^t(\utraj_R,\model)) \enspace,
    \label{eq:R_solution}
\end{equation}
where $R_R$ is evaluated only over the first $K$ states starting at $\state^t$.
The robot executes the first action from its plan and then replans at the next time step $t+1$. To simplify notation, we will denote the first action of a plan as $\planControlAct^t(\model) \coloneqq  \planAct^{t}(\model)[0]$. 

The crucial question is what human model $\model$ should the robot use for planning with Eq. \eqref{eq:R_solution}? Restricting ourselves to any single model either hurts performance or computational efficiency.  We propose an algorithm which enables efficient switching between models of varying accuracy and complexity, allowing for both high performance and low computation.



\subsection{Model Switching Formalism}

We assume the robot has access to a ladder of human models $\mathcal{M} = \{\model_0, \hdots, \model_n\}$, where as we climb the ladder we sacrifice computational time for greater expected reward: 
$\mathbb{E}_{x}[R_R(\state,\planAct(\model_j), \utraj_H)] \geq \mathbb{E}_{x}[R_R(\state,\planAct(\model_i), \utraj_H)]$ and $T(\model_j) > T(\model_i), \forall i < j$, where $T(\model)$ is the time to solve Eq. \eqref{eq:R_solution} under $\model$.

At every time step $t$, the robot needs to choose the human model $\model^t\in \mathcal{M}$ to use for planning $\planAct^t$.
To determine which model the robot should use at time $t$, we construct a \emph{meta MDP} on top of the previous MDP, with states $x^t$, actions $\model^t\in\mathcal{M}$ representing the choice of human model, transition function $f(\state^t, \planControlAct^t(\model^t), \control_H^t)$, and meta-reward $r_{meta}^t = r_R(\state^t, \planControlAct^t(\model^t), \control_H^t) - \lambda*T(\model^t)$, with $\lambda$ trading off actual reward gained by planning under $\model^t$ and computational time spent on the plan. 
Lower values for $\lambda$ favor more complex models, whereas higher values result in more usage of less expensive ones.

Solving this MDP exactly is impossible, since it requires access to the true human controls $\utraj_H$ ahead of time. Moreover, even if we approximate the true human controls with those predicted by our most accurate model $M_n$, the robot needs to find the model sequence that maximizes the cumulative meta reward across the episode---a procedure exponential in the episode horizon and, thus, intractable.
We could alleviate this computational burden by assuming the robot myopically decides when to switch models: instead of considering the cumulative meta reward, only look at $r_{meta}^t$ to decide whether to switch at time $t+1$. On the plus side, this simplification is more tractable and only needs the current human control. Unfortunately, even this relaxation requires planning with every model and picking the one with the best $r_{meta}^t$,
which is worse than simply using $\model_n$ from the get-go. Thus, we now propose an approximate solution that avoids computing every plan while still switching models as needed.

\subsection{Approximate Solution: Switching between Two Models}
\label{sec:2model_solution}


For ease of exposition, we first discuss how the robot can decide whether to switch from a simple model $\model_1$, used for planning at timestep $t$, to a complex model $\model_2$, which we are considering for timestep $t+1$. We are interested in estimating the reward the robot would gain from using $\model_2$.


\noindent\textbf{Estimate Change in Robot Plan}

Leveraging our plan computed at timestep $t$ using $\model_1$, we get around explicitly generating $\planAct^t(\model_2)$ by approximating it as $\plan^t(\model_2) = \planAct^t(\model_1) + \Delta \plan$. 
Here, we want to choose $\Delta \plan$ such that it maximizes the robot reward $R_R$ under the complex model $\model_2$.
However, optimizing $\Delta \plan$ using Eq.~\eqref{eq:R_solution} is equivalent to planning. 
To get an efficient estimate, we use a quadratic Taylor series approximation of $R_R$, denoted $\tilde{R}_R$, evaluated around $(\state^t, \planAct^t(\model_1), \modelTrajAct^t(\planAct^t(\model_1), \model_1))$, our current plan and human prediction:
%
\begin{equation}
\Delta\plan = 
\arg\max_{\Delta\utraj_R}
 \tilde{R}_R(\state^t,\plan^t(\model_2), \modelTrajAct^t(\plan^t(\model_2), \model_2))\enspace.
\label{eq:delta_R_optimization}
\end{equation}
%

\noindent\textbf{Estimate Change in Human Prediction}

Eq.~\eqref{eq:delta_R_optimization} requires approximating the $\model_2$ human's response $\modelTrajAct^t$ to the robot's plan $\plan^t(\model_2)$, which can be broken down into $\model_2$'s response to the current plan $\planAct^t(\model_1)$ plus the change coming from $\Delta\plan$:
$\modelTraj^t(\plan^t(\model_2), \model_2) = \modelTraj^t(\planAct^t(\model_1) + \Delta\plan, \model_2)$.
We can linearly approximate this:
\begin{equation}
    \modelTraj^t(\plan^t(\model_2), \model_2) = \modelTrajAct^t(\planAct^t(\model_1), \model_2) + \frac{d\modelTraj}{d\plan}\cdot \Delta\plan\enspace,
    \label{eq:human_control_under_j}
\end{equation}
where $\frac{d\modelTraj}{d\plan} = \frac{d\modelTraj(\plan,\model)}{d\plan}\Bigr|_{\substack[b]{ \plan=\planAct^t(\model_1)\\ \model=\model_2}}$. This requires evaluating the complex model $\model_2$, which might be expensive, though not nearly as expensive as planning with it.


We may now substitute the changed robot plan $\plan^t(\model_2)$ and the changed human prediction from Eq.~\eqref{eq:human_control_under_j} into our reward expression we maximize in Eq.~\eqref{eq:delta_R_optimization}
%
\begin{equation}
\tilde{R}_R(\state^t,\planAct^t(\model_1) + \Delta\plan, \modelTrajAct^t(\planAct^t(\model_1), \model_2) + \frac{d\modelTraj}{d\plan}\cdot \Delta\plan)\enspace.
\label{eq:taylor}
\end{equation}


\noindent\textbf{Performance Gain from the Change in Robot Plan}

Ultimately, to assess the robot's performance gain by using $\model_2$, we want to estimate the reward component of $r_{meta}^t$ under $\model_2$, $r_R(\state^t, \planControl^t(\model_2), \modelControlAct^t(\plan^t(\model_2), \model_H))$, where $\modelTrajAct^t(\plan^t(\model_2), \model_H)$ would be the human's response under the true model $\model_H$ to the robot's plan with $\model_2$, and $\modelControlAct^t(\plan^t(\model_2), \model_H)$ is the first control of that response.
To simplify notation, we will denote this reward $r_R^t(\model_2)$.

Since our approximation of the meta MDP considers myopic rewards $r_R$, we really are only interested in the first control $\planControl^t(\model_2)$ of the changed plan $\plan^t(\model_2)$, and the first control $\modelControlAct^t(\plan^t(\model_2), \model_H)$ of the true human's response to that plan. 
Given $\Delta\plan$, we only need the change for the first control $\Delta\planControl$ combined with the approximation of $\modelControlAct^t(\plan^t(\model_2), \model_H)$ as in Eq.~\eqref{eq:human_control_under_j} to estimate $\hat{r}_R^t(\model_2)$ as:
%
\begin{equation}
    r_R(\state^t, \planControlAct^t(\model_1) + \Delta\planControl, \modelControlAct^t(\planAct^t(\model_1), \model_H) + \frac{d\modelControl}{d\planControl} \cdot\Delta\planControl)\enspace,
    \label{eq:R_j_estimate}
\end{equation}
where $\frac{d\modelControl}{d\planControl}=\frac{d\modelControl(\plan,\model)}{d\planControl}\Bigr|_{\substack[b]{\plan=\planAct^t(\model_1)\\ \model=\model_H}}$. Here, we don't know $\model_H$, but we can approximate it with the complex model $\model_2$.

\noindent\textbf{One Time Step Simplification}

All of this requires evaluating the complex model $\model_2$ at the current plan $\planAct^t(\model_1)$, which, although cheaper than fully planning with $\model_2$, is still expensive. Since the meta reward only evaluates the first control of the plans, we can do better by simplifying our formulation further to planning and predicting a single control. That is, we consider the changed control $\planControl^t(\model_2)=\planControlAct^t(\model_1)+\Delta\planControl$, with its corresponding changed human control prediction from Eq.~\eqref{eq:human_control_under_j}:
\begin{equation}
    \modelControl^t(\plan^t(\model_2), \model_2) = \modelControlAct^t(\planAct^t(\model_1), \model_2) + \frac{d\modelControl}{d\planControl}\cdot \Delta\planControl \enspace,
    \label{eq:human_control_approx}
\end{equation}
where $\frac{d\modelControl}{d\planControl} = \frac{d\modelControl(\plan,\model)}{d\planControl}\Bigr|_{\substack[b]{ \plan=\planAct^t(\model_1)\\ \model=\model_2}}$.

We obtain $\Delta\planControl$ by optimizing the following objective:
\begin{equation}
\arg\max_{\Delta\control_R}
 \tilde{r}_R(\state^t,\planControl^t(\model_2), \modelControlAct^t(\planAct^t(\model_1), \model_2) + \frac{d\modelControl}{d\planControl}\cdot \Delta\control_R)
\enspace.
\label{eq:delta_control}
\end{equation}
%
Our simplified derivation still requires the gradient $\frac{d\modelControl}{d\planControl}$, but this is cheaper to compute than $\frac{d\modelTraj}{d\plan}$.


Armed with an estimated $\hat{r}_R^t(\model_2)$ from Eq.~\eqref{eq:R_j_estimate}, we can now determine whether the performance gain from $\model_2$ is worth the additional compute by considering
\begin{equation}
    \Delta r^t_{meta} = \hat{r}_R^t(\model_2) - \lambda*T(\model_2) - (r_R^t(\model_1) - \lambda*T(\model_1))\enspace,
    \label{eq:delta_r_meta}
\end{equation}
where we already know $r_R^t(\model_1)$ from the previous time step, and $T(\model_1)$ and $T(\model_2)$ are known a priori from their model specifications. If $\Delta r^t_{meta}$ is positive, the robot should switch to $\model_2$; otherwise, the robot should continue using $\model_1$.

\noindent\textbf{Intuitive Interpretation}

In Eq.~\eqref{eq:delta_r_meta}, since $T$'s are constants, the robot's decision of whether to switch relies on comparing $r_R^t(\model_1)=r_R(\state^t, \planControlAct^t(\model_1), \modelControlAct^t(\planAct^t(\model_1), \model_1))$ to $\hat{r}_R^t(\model_2)=r_R(\state^t, \planControlAct^t(\model_1) + \Delta\planControl, \modelControlAct^t(\planAct^t(\model_1), \model_2) + \frac{d\modelControl}{d\planControl} \cdot\Delta\planControl)$. Looking at these rewards, a few key distinctions stand out.

First, the robot is interested in knowing whether $\model_2$ would give a different prediction $\modelControlAct^t(\planAct^t(\model_1), \model_2)$ than $\model_1$'s $\modelControlAct^t(\planAct^t(\model_1), \model_1)$ on the current plan. For example, in Fig. \ref{fig:front_fig} we realize a more complex model foresees the human turning into the bottleneck compared to naive constant velocity. Here this foresight prevents a collision.


Meanwhile, the derivative $\frac{d\modelControl}{d\planControl}$ captures the \textit{influence} that the robot's controls have on the human's. Simple models may ignore this, but more complex models, like the two player game, capture this dynamic enabling certain critical maneuvers. For instance, humans will yield should the robot merge to create proper spacing. Only if aware of this influence will the robot feel confident merging into tight windows.


So how much do these two terms matter? Intuitively, by computing $\Delta\planControl$ via Eq.~\eqref{eq:delta_control}, we see how much a model that either captures influence or makes different predictions affects the robot's plan. If neither bears much weight, then $\Delta\planControl$ will likely be negligible and we will not switch. If however $\Delta\planControl$ is significant, we further evaluate if that translates to significant performance gain. Only when that's the case will we ultimately switch.

\begin{algorithm}[t]
\DontPrintSemicolon
\textbf{Input:} Ladder $\mathcal{M} = \{\model_0, \hdots, \model_n\}$, episode time $N$.\\
Start with time $t=0$, current model index $i$.\\
\While{$t \leq N$}{
    Compute $\planAct^{t}(\model_i)$ given $x^t$ and execute $\planControlAct^{t}(\model_i)$.\\
    \If{$i < n$}{
    Substitute $(\model_1, \model_2) \gets (\model_i, \model_n)$ in Eq.~\eqref{eq:R_j_estimate}, \eqref{eq:delta_control}, \eqref{eq:delta_r_meta}, with $\modelControlAct^t(\planAct^t(\model_i), \model_n) \gets \control_H^t$. \\
    Compute $\Delta r^t_{meta}$ using Eq.~\eqref{eq:R_j_estimate}, \eqref{eq:delta_control}, \eqref{eq:delta_r_meta}.\\
    \If{$\Delta r^t_{meta}>0$}{
    $i \gets n$ (Switch up), continue\\
    }}
    \If{$i > 0$ and cooldown complete}{
    Substitute $(\model_1, \model_2) \gets (\model_i, \model_{i - 1})$ in Eq. ~\eqref{eq:R_j_estimate}, \eqref{eq:delta_control}, \eqref{eq:delta_r_meta}. \\
    Compute $\Delta r^t_{meta}$ using Eq.~\eqref{eq:R_j_estimate}, \eqref{eq:delta_control}, \eqref{eq:delta_r_meta}. \\
    \If{$\Delta r^t_{meta}>0$}{
    $i\gets i-1$ (Switch Down).\\
    }
    }
}
\caption{Dynamic Model Switching}
\label{alg:model_switching}
\end{algorithm}

\subsection{Approximate Solution: Switching Between the Ladder}

Although we presented our method in the context of switching up, the framework holds when $\model_1$ is the current model and $\model_2$ is any alternative: higher or lower. When the alternate model is lower the question becomes: is the computational saving worth the loss in reward? Generalizing our derivation to a ladder of models $\mathcal{M} = \{\model_0, \hdots, \model_n\}$, suppose that at time $t$ the robot used model $\model_i$ to plan $\planAct^t(\model_i)$. We would like to decide on a model for time $t+1$.


First, we evaluate if it's worthwhile to switch to a higher model $\model_j$, $j > i$. The robot could consider $\model_{i+1}$, the model immediately above, and successively switch up. However, in urgent, safety-critical situations we may need to switch higher than that immediately to avoid an accident. Thus, we exclusively consider the best model and upper bound $\hat{r}_R^t(\model_j)$ with $\hat{r}_R^t(\model_n)$, the estimated reward from using the best model available. To avoid having to evaluate $\model_n$'s expensive predictions we substitute a perfect prediction $\modelControlAct^t(\planAct^t(\model_i), \model_N) \approx \control_H^t$.  This effectively upper bounds $\model_n$ with the true human model $M_H$.

If the robot should have switched down to $\model_j$, $j<i$, we observe that $r_R^t(\model_j) \leq  r_R^t(\model_{i-1}) \leq r_R^t(\model_i)$. Thus, for efficient switching, we only consider the model directly below, $\model_{i-1}$. Since $T(\model_{i-1}) < T(\model_{i})$, it is reasonable to compute true model predictions in our approximation. 

Finally, if $\Delta r^t_{meta}$ is positive for $\model_H$, the robot switches up to $\model_n$. Otherwise, if $\Delta r^t_{meta}$ is positive for $\model_{i-1}$, the robot switches down to $\model_{i-1}$. Otherwise we stay as summarized in Algorithm \ref{alg:model_switching}.


In practice evaluating $\model_{i-1}$'s predictions can still be significant, but unlike switching up, safety and performance concerns do not force us to check every timestep. One may wait $K$ timesteps after failing to switch down before trying again. This cooldown hyperparameter, $K$, should be set based on how often one actually switches.

%% file: 3_experiments.tex
\section{Experiments}
\label{sec:experiments}

\begin{figure*}
\centering
\begin{subfigure}{.3\textwidth}
  \centering
  \includegraphics[width=\textwidth,clip=false]{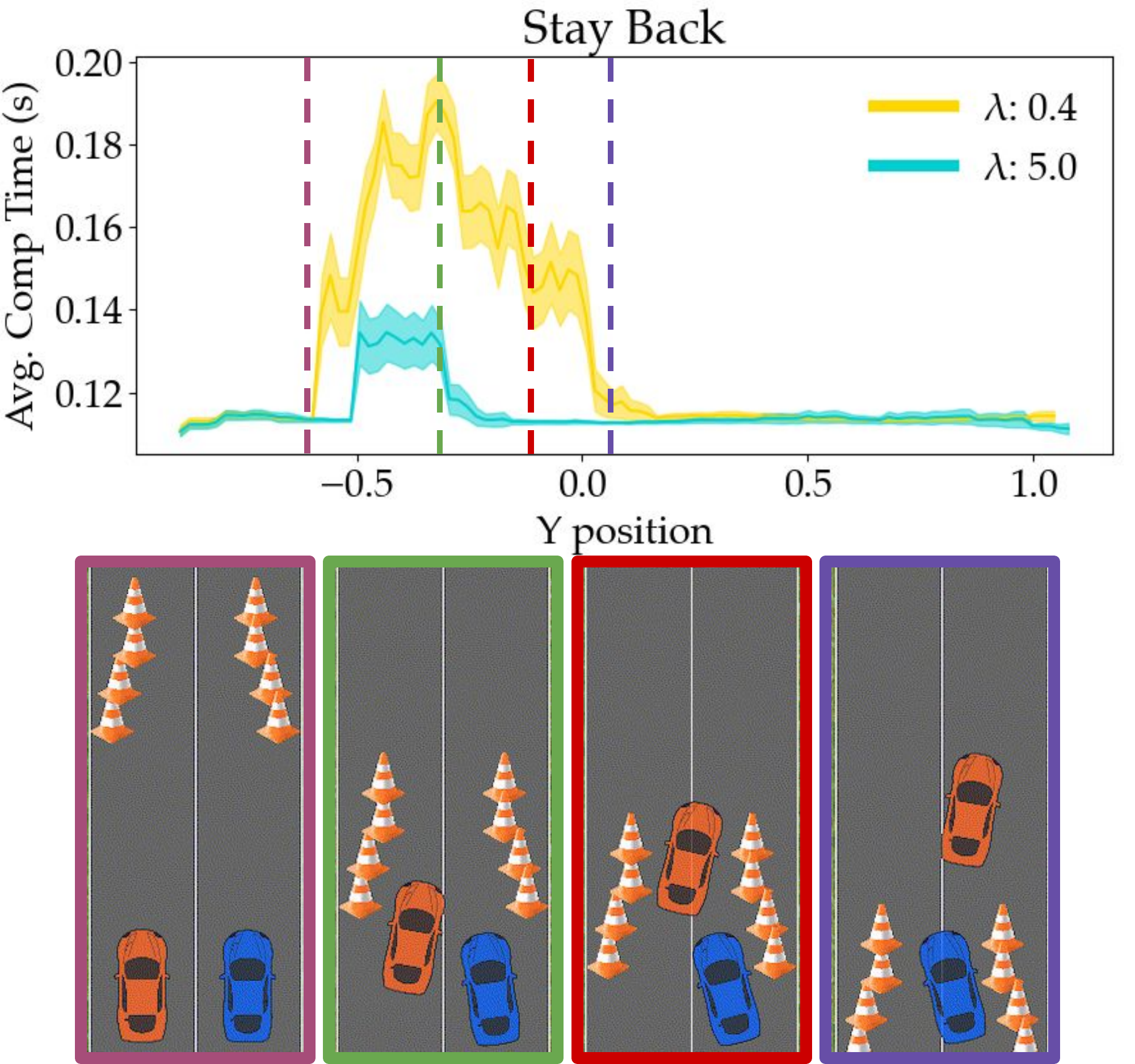}
\end{subfigure}
\centering
\begin{subfigure}{.3\textwidth}
  \centering
  \includegraphics[width=\textwidth,clip=false]{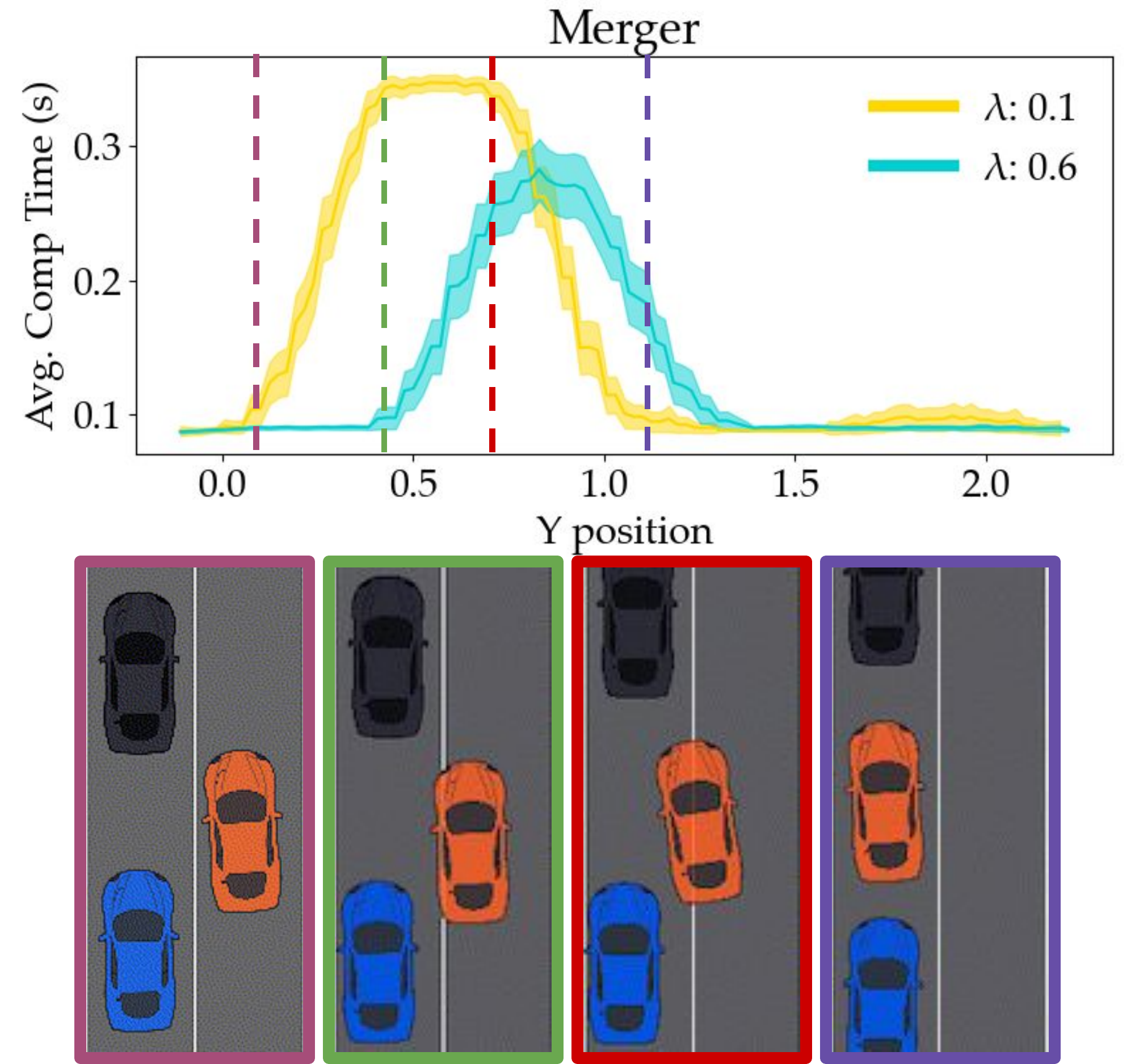}
\end{subfigure}
\centering
\begin{subfigure}{.3\textwidth}
  \centering
  \includegraphics[width=\textwidth]{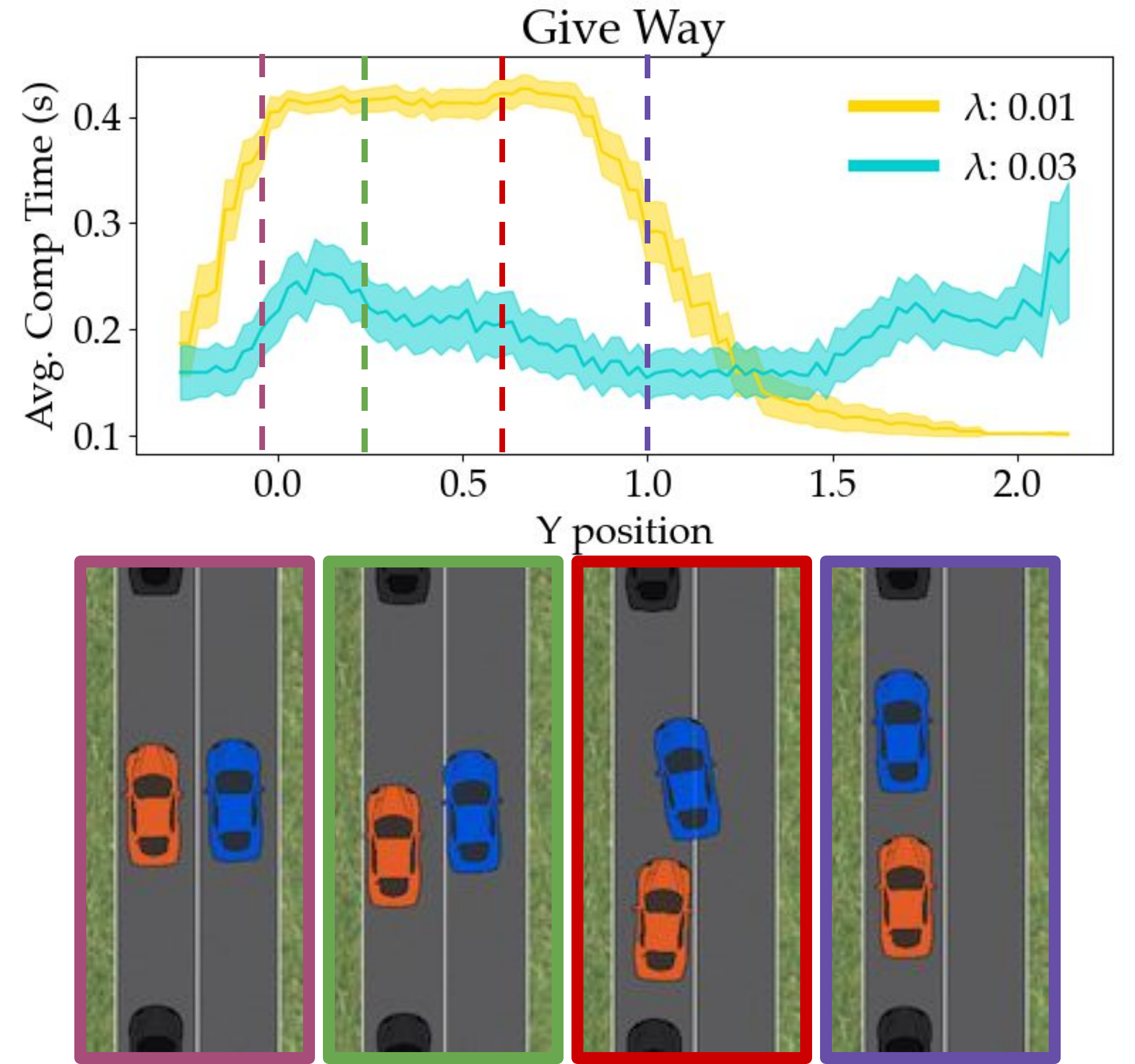}
\end{subfigure}
\caption{(Top) Average single step computation time over the course of Stay Back (left), Merger (center), and Give Way (right), for a conservative, lesser $\lambda$, switcher (yellow) and an aggressive, larger $\lambda$, one (light blue). (Bottom) Example conservative MS robot (orange) behavior around the target human (blue) and other cars (black).}
\label{fig:qualitative}
\end{figure*}

We now demonstrate the efficacy of our model switching algorithm in three simulated autonomous driving scenarios.

\subsection{Driving Simulator}

We model the dynamics of the vehicles as a 4D bicycle model. Let the state of the system be $\xtraj = [x \; y \; \theta \; v]^T$, where $x, y$ are the coordinates of the vehicle, $\theta$ is the heading, and $v$ the speed. The actions are $\control = [\omega \; a]^T$, where $\omega$ is the steering input and $a$ is the linear acceleration. We use $\alpha$ as the friction coefficient, and the vehicle's dynamics model is:
\begin{equation}
    [\dot{x} \; \dot{y} \; \dot\theta \; \dot{v}] = [v\cdot\cos(\theta) \;\;\; v\cdot\sin(\theta) \;\;\; v\cdot\omega \;\;\; a - \alpha\cdot v] \enspace.
\end{equation}

\subsection{Human Predictive Models} 

For our experiments, we use the following 3 models of varying computational complexity and accuracy:

\subsubsection{Constant Velocity} This model, deemed \textit{Naive}, predicts that the person will provide zero acceleration control, maintaining their current heading and speed.

\subsubsection{Human Plans First} This model assumes that the person acts according to a reward function parameterized as a linear combination of features $\phi$: $R_H(\state^0,\utraj_R,\utraj_H) = \sum_{\tau=1}^N r_H(\state^\tau, \control_R^\tau, \control_H^\tau) = \sum_{\tau=1}^N \theta^T \phi(\state^\tau, \control_R^\tau, \control_H^\tau)$. 
Additionally, under this model the human believes that the robot will maintain a constant velocity, and optimizes their reward w.r.t the imagined robot plan $\tilde\utraj_R$ to obtain a plan $\modelTrajAct$:
\begin{equation}
    \modelTrajAct(\state^0, \tilde\utraj_R) = \arg\max_{\utraj_H} R_H(\state^0, \tilde\utraj_R, \utraj_H)\enspace.
\end{equation}
We refer to this model as \textit{Turn} because the person takes the first turn in choosing their controls.

\subsubsection{Cognizant of Effects on Human Action} Our last model is based on~\cite{Sadigh2016PlanningFA} and models the human as an agent which will optimally respond to the robot's plan. This results in a nested optimization as the robot accounts for how its plan affects the human's. As we rationalize the human's thought process we refer to this model as ``Theory of Mind''~\cite{gopnik_wellman_1994}, or \textit{ToM} for short.

Planning with the first two approaches involves optimizing the robot's control against the fixed human prediction. Meanwhile, ToM's nested optimization returns both a human and robot control with no further optimization required. From our experiments, planning using Turn and ToM takes roughly twice and four times as long as using Naive, respectively. However, their expected accuracy has the reverse ordering.

Both Turn and ToM rely on knowing a reward parameter $\theta$ and a set of features $\phi$ for the human reward. To learn a good $\theta$ for every scenario, we collected demonstrations of a single human driver in an environment with multiple autonomous cars following precomputed routes, and performed inverse reinforcement learning~\cite{maxent}. The base features we used include higher speed moving forward (against a speed limit), lateral and directional alignment in the lane, collision avoidance, and distance to the boundaries of the road. 
%
%

Across our experiments, we compare each of these three models against our model switcher (MS) that dynamically chooses between all of them. We additionally wanted to analyze different performances for designers that might be more or less conservative about the reward vs. computation tradeoff, so we show results for different values of $\lambda$.

\subsection{Miscellaneous Experimental Details}
We conduct experiments using TensorFlow~\cite{tensorflow2015-whitepaper} 2.1, running on a 2015 MacBook Pro, for gradient calculation and optimization.  All planners optimize for a horizon $T = 5$ using 20 vanilla gradient descent steps.  For switching down, we used a cooldown of $K = 3$.
In Eq. \eqref{eq:delta_R_optimization}, the quadratic approximation may be ill conditioned, so we restricted $\Delta\control_R$ to exist within some reasonable bounds.

\subsection{Evaluation Strategy}

For every scenario, we run every method against the same simulated human driver. For diversity, we vary the starting positions of the human and robot cars across 30 different seeds per scenario. For every time step, we keep track of the reward and computation time.

In Fig. \ref{fig:qualitative} we visualize average combined planning and decision time for the MS as it progresses through each scenario.  Further, we juxtapose a conservative designer (yellow), who values reward relatively more, with an aggressive one (blue), who prefers computational savings.  Snapshots below provide context for key points of the experiment denoted by dashed lines above, taken from a single conservative MS run.  

\begin{figure*}
\centering
\begin{subfigure}{.32\textwidth}
  \centering
  \includegraphics[width=\textwidth,clip=false]{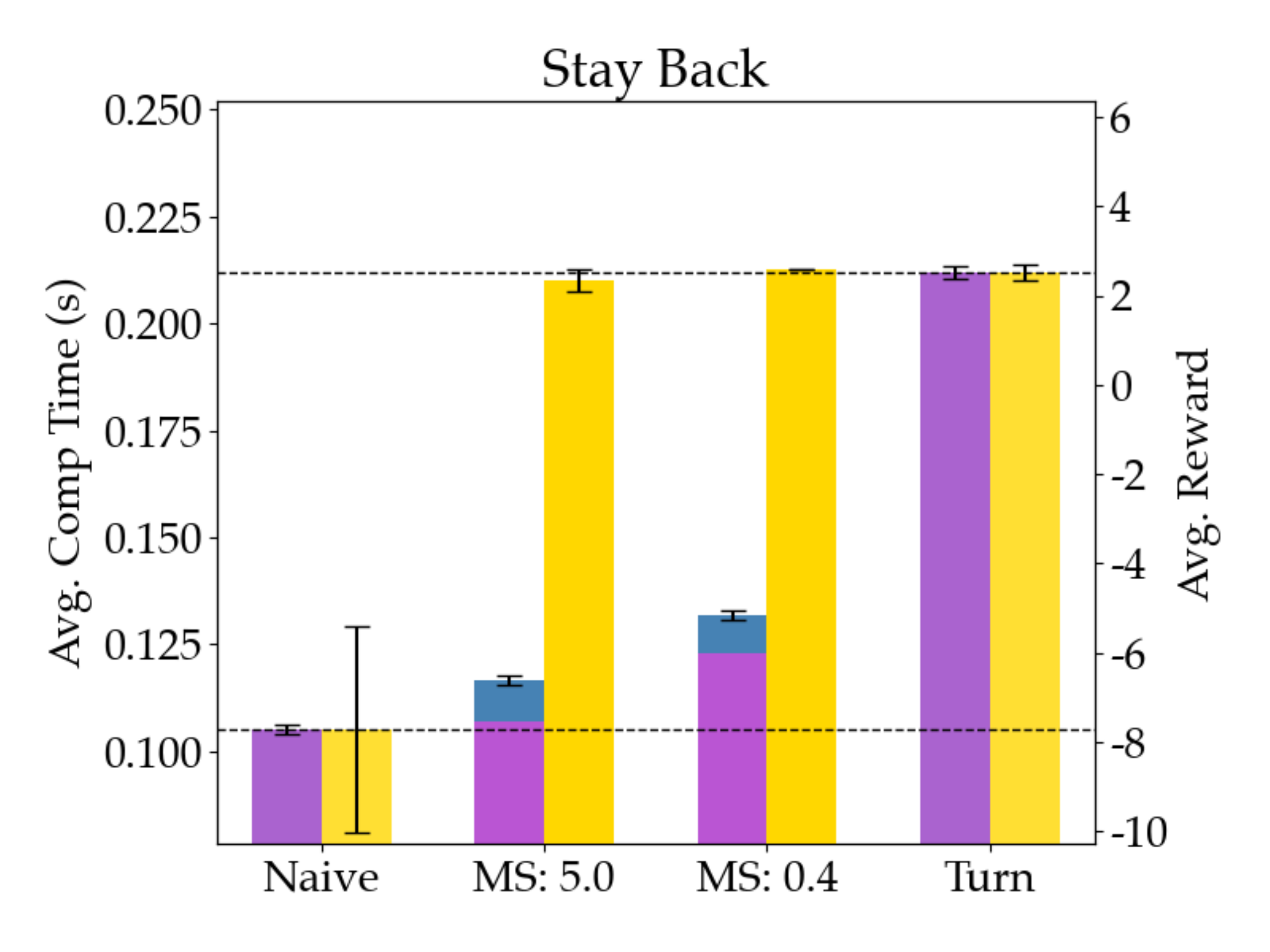}
\end{subfigure}
\centering
\begin{subfigure}{.33\textwidth}
  \centering
  \includegraphics[width=\textwidth,clip=false]{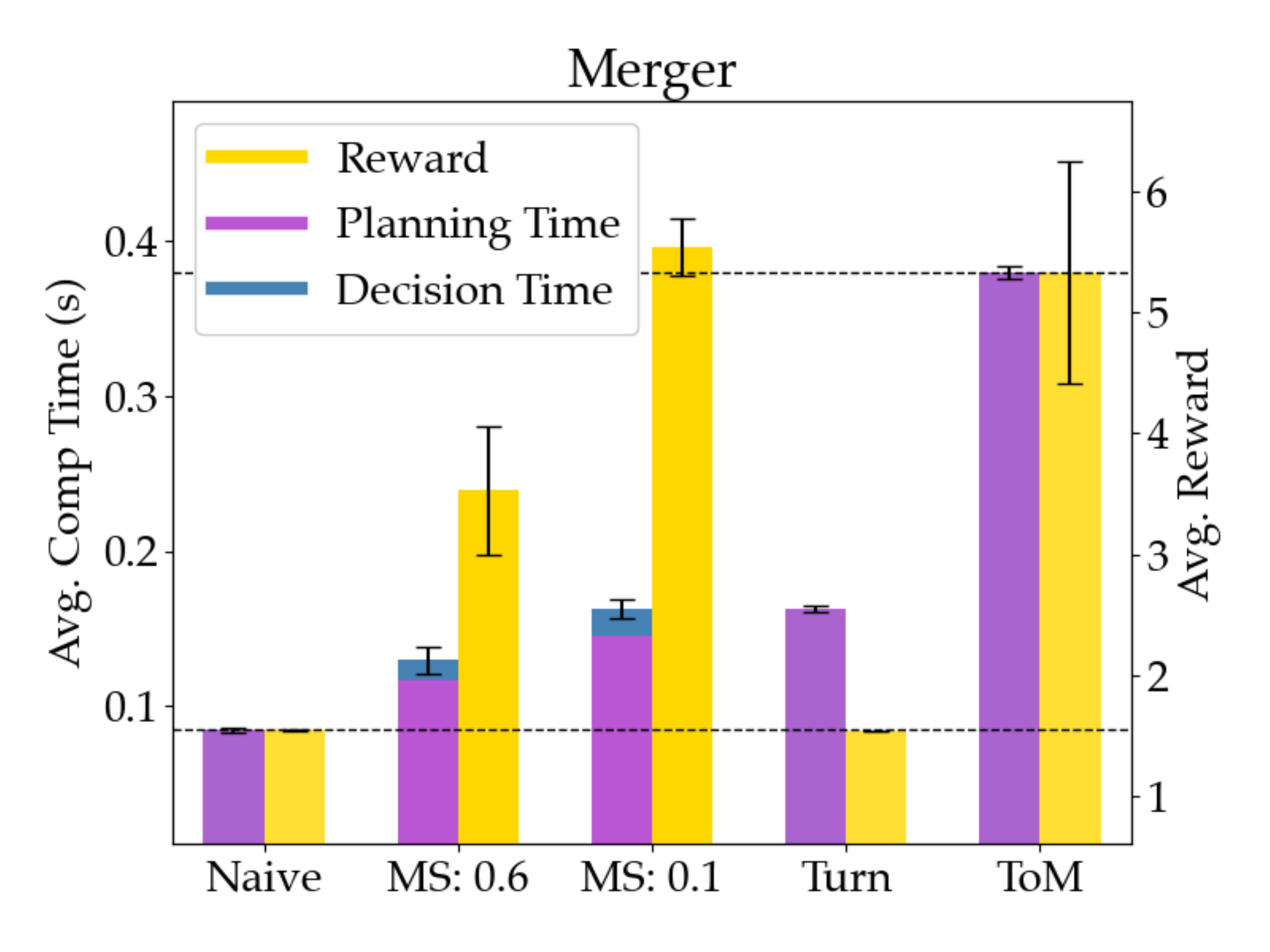}
\end{subfigure}
\centering
\begin{subfigure}{.33\textwidth}
  \centering
  \includegraphics[width=\textwidth]{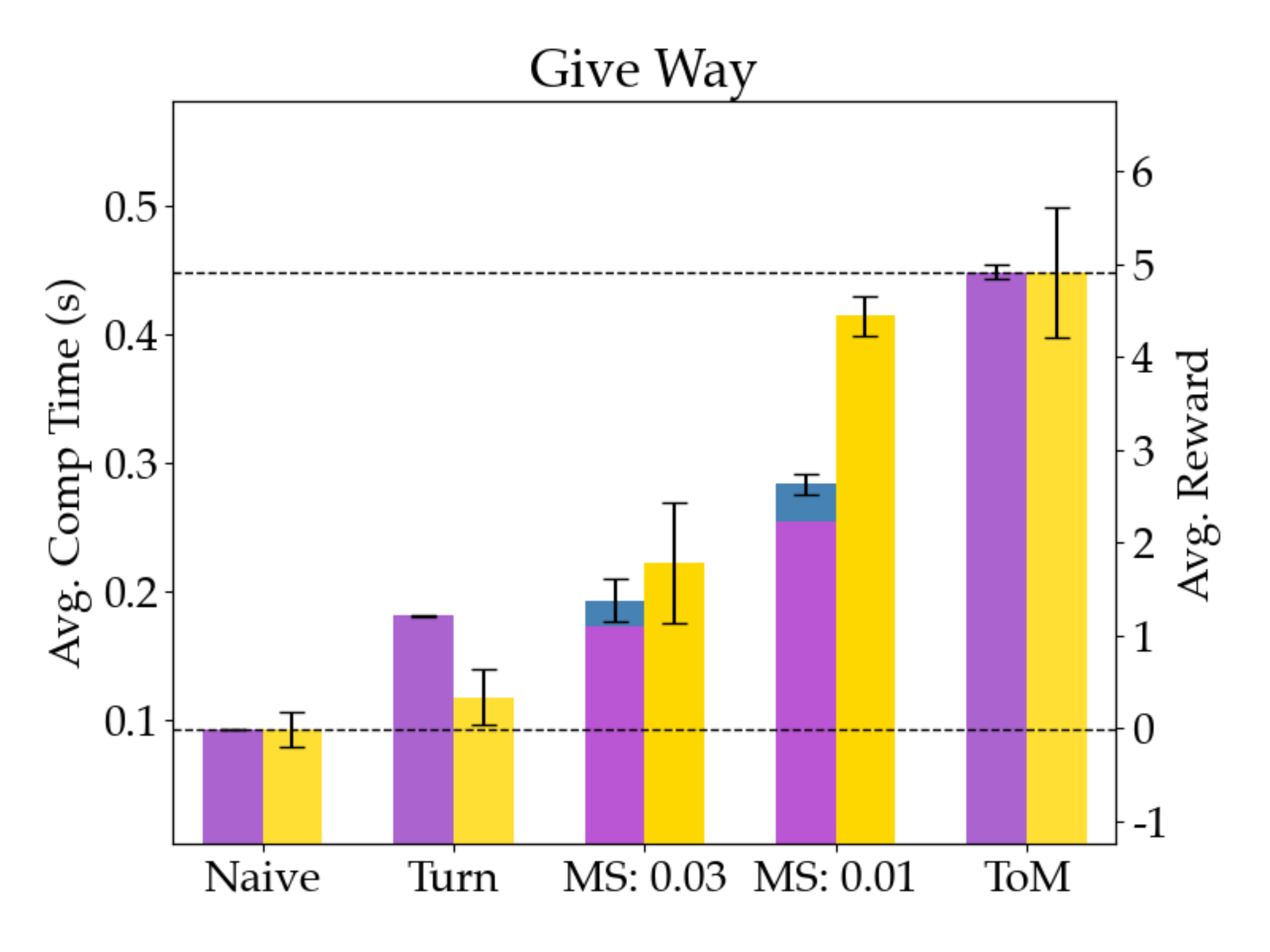}
\end{subfigure}
\caption{Average computational time and reward for the 3 scenarios, with models ordered by computation. Model switchers achieve comparable performance to the best model with less computation.  Note: the aggressive switchers with greater values for $\lambda$ are to the left of the conservative ones.}
\label{fig:quantitative}
\end{figure*}

In Fig. \ref{fig:quantitative}, we showcase the average reward and computational time per episode, averaged across seeds. In each plot, the left bars represent computational time, and the right ones reward, in units given by the left and right axes respectively. Within the total computational time, we separate planning and time deciding a human model.

We hypothesized that our MS algorithm would maintain a reward similar to that of the top model, while being significantly cheaper to compute. Additionally, we expected to see that conservative switchers obtain better rewards than aggressive switchers, but at higher computational complexity.

Note that because we wanted to showcase regions where a conservative switcher would react differently from an aggressive one, the hyperparameter $\lambda$ varies widely across our scenarios. This is a reflection of differing reward scales, and in general the designer would have to select an appropriate $\lambda$ depending on the problem and desired reward-computation tradeoff. For greater stability and generalization, one may find a highly conservative $\lambda$ to be effective, reducing switching sensitivity to situations where the human has little bearing on the reward.

\subsection{Scenario 1: Stay Back}
In our first scenario in Fig. \ref{fig:qualitative} (left), the robot and human begin driving alongside each other at the same speed. Ahead a series of cones creates a bottleneck in the road. Either the robot or human must yield to avoid a collision.

For this scenario, we restrict ourselves to Naive and Turn for simplicity, while the other two will showcase the potential of a broader model ladder. 
As shown in Fig. \ref{fig:quantitative} (left), the Naive model struggles to correctly anticipate whether the human will go first and act accordingly, but it is much cheaper than Turn which safely navigates the scenario.
Meanwhile, both aggressive or conservative switchers manages to obtain rewards close to that of the Turn, but with computational time closer to Naive. Additionally, notice that the decision time doesn't add excessive overhead, which underlines the efficiency of our approximation to the meta MDP.

In Fig. \ref{fig:qualitative} (left), we see that a conservative switcher ($\lambda = 0.4$) will generally use Turn before and during the bottleneck, whereas an aggressive one ($\lambda = 5.0$) uses Turn only to intervene when using Naive is headed towards a collision.

\subsection{Scenario 2: Merger}
In the next scenario shown in Fig. \ref{fig:qualitative} (middle), the robot would like to merge into the left lane. The gap is too small for the robot, but if it gradually angles its hood the target human will yield allowing the robot to enter.

As shown in Fig. \ref{fig:quantitative} (middle), only ToM is able to anticipate that the human will yield should it begin entering the lane. However, before and after merging, ToM provides no further advantage, so the switcher can exploit that to reduce computational complexity.

In Fig. \ref{fig:quantitative} (middle), we see that both the conservative and aggressive switchers obtain rewards closer to ToM, but with significantly less total computation. In Fig. \ref{fig:qualitative} (middle), we see that the model switcher only uses ToM for merging, as it provides no comparative advantage afterwards. A conservative model switcher ($\lambda = 0.1$) switches up earlier merging faster, whereas an aggressive one ($\lambda = 0.6$) switches later but requires less computation.  Delaying an inevitable switch hurts the aggressive MS, highlighting the importance of a conservative $\lambda$ for safety-critical applications Turn is used sparingly during the transition as it provides little comparative advantage to Naive.

\subsection{Scenario 3: Give Way}
In our last scenario shown in Fig. \ref{fig:qualitative} (right), the person is driving alongside the robot and would like to enter the robot's lane; however, other drivers around the robot do not allow enough space either way. The robot would like to help the human enter. Of course, the robot may create sufficient space by moving forward or backing up.

As shown in Fig. \ref{fig:quantitative} (right), ToM is the only one capable of understanding the robot's ability to help the person to merge.  Turn occasionally succeeds when it yields fearing a collision. Both switchers obtain higher rewards than those of the cheaper models, but we notice again a large gap between the two.

The earlier the robot makes space for the person, the better. A conservative model switcher ($\lambda = 0.01$) quickly switches up and allows the human to enter, albeit slower than pure ToM. A more aggressive switcher ($\lambda = 0.03$) delays the switch to ToM resulting in lesser reward, but keeps overall computation lower. The delay between yielding and the human entering presents a challenge for our myopic, one time step, reward simplification. A very low $\lambda$ works here, but the issue of myopic gain potentially underestimating longer horizon gain remains.

%% file: 4_conclusion.tex
\section{Discussion}
\label{conclusion}

\textbf{Summary:} In this paper, we formalized the robot's decision making process over which predictive human model to use as a meta MDP. We introduced an approximate solution that enables efficient switching to the most suitable available model within this MDP. The resulting decisions maintain rewards similar to those of the best model available, while dramatically reducing computational time.

\textbf{Future Work:} Because the robot cannot see the human's true future controls, we were limited to basing switching decisions on what the person did in the past. We could approximate the true human trajectory using the top model's prediction, but that would relinquish most computational savings.  For example, Naive planning and Turn prediction takes as long as Turn planning.  Additionally, because the decision to switch relies on a single time step simplification, our scenarios needed consistent reward signal. Future work must address adapting our algorithm to sparse reward settings where one step reward gradients are not meaningful. Learning a value function to replace reward in our formulation would be an interesting direction.

Moreover, all this work happened in a simple driving simulator, albeit with what we think are complex scenarios. To put this on the road, we will need more emphasis on safety, as well as longer decision horizons. Lastly, our algorithm focuses on single human decisions. We could run our method separately for every nearby human, evaluating the differential benefit of switching each, but we are yet to conduct experiments in that setting.  Alternatively, we can imagine adapting it to multiple humans by either adding more complex multi-player game theoretic models, or combining it with the prioritization schema presented by \cite{refaat2019prioritization}.

\textbf{Conclusion:} Despite these limitations, we are encouraged to see robots have more autonomy over what human models to use when planning online, without hand-coded heuristics. We look forward to applications of our model switching ideas beyond autonomous driving: to mobile robots, quadcopters, or any human-robot interactive scenarios where planning with multiple predictive human models might be beneficial.